\pdfoutput=1

\documentclass[11pt]{article}

\usepackage{ACL2023}

\usepackage{times}
\usepackage{latexsym}

\usepackage[T1]{fontenc}

\usepackage[utf8]{inputenc}
\usepackage{graphicx}

\usepackage{microtype}

\usepackage{inconsolata}
\usepackage{amsmath}
\usepackage{fix-cm}

\title{Efficient Zero-Shot Long Document Classification by Reducing Context Through Sentence Ranking}

\author{
    Prathamesh Kokate\textsuperscript{1,3}, Mitali Sarnaik\textsuperscript{1,3}, Manavi Khopade\textsuperscript{1,3}, Mukta Takalikar\textsuperscript{1}, and Raviraj Joshi\textsuperscript{2,3}\thanks{Correspondence to: ravirajoshi@gmail.com} \\
    \textsuperscript{1}Pune Institute of Computer Technology, Pune \\
    \textsuperscript{2}Indian Institute of Technology Madras, Chennai \\
    \textsuperscript{3}L3Cube Labs, Pune
}

\begin{document}
\maketitle
\begin{abstract}
Transformer-based models like BERT excel at short text classification but struggle with long document classification (LDC) due to input length limitations and computational inefficiencies. In this work, we propose an efficient, zero-shot approach to LDC that leverages sentence ranking to reduce input context without altering the model architecture. Our method enables the adaptation of models trained on short texts, such as headlines, to long-form documents by selecting the most informative sentences using a TF-IDF-based ranking strategy. Using the MahaNews dataset of long Marathi news articles, we evaluate three context reduction strategies that prioritize essential content while preserving classification accuracy. Our results show that retaining only the top 50\% ranked sentences maintains performance comparable to full-document inference while reducing inference time by up to 35\%. This demonstrates that sentence ranking is a simple yet effective technique for scalable and efficient zero-shot LDC.
\end{abstract}

\section{Introduction}
\begin{figure*}
  \centering
  \includegraphics[width=\textwidth]{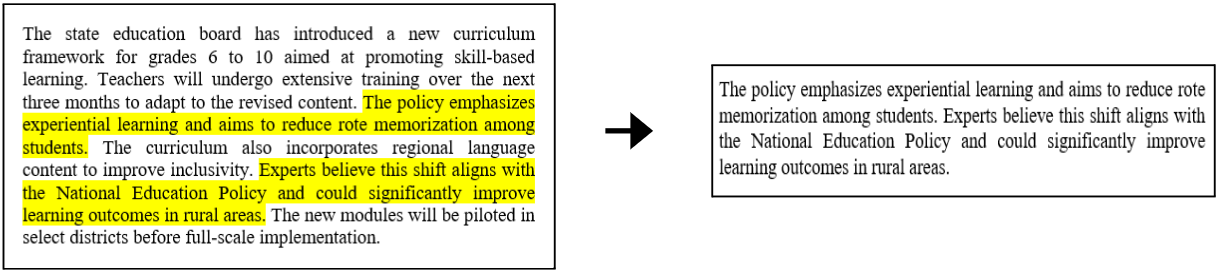}
  \caption{ Long-to-Short Document Transformation via Sentence Ranking. The figure shows how our method converts a long document (left) into a shorter version (right) by selecting top-ranked sentences using TF-IDF scores. These highlighted sentences capture the core meaning of the document, enabling a model trained on short texts to classify long documents accurately with reduced input and faster processing.
}
  \label{fig:paragraph}
\end{figure*}

Transformer-based language models such as BERT have revolutionized short text classification, delivering impressive results on tasks like short headline classification (SHC) or tweet categorization. However, their effectiveness diminishes when applied to long document classification (LDC), primarily due to fixed input length constraints and increased computational overhead. These challenges are further amplified in real-world deployments, where long texts are common and efficiency is critical.
In this work, we target the problem of efficient zero-shot long document classification, where models trained only on SHC data are applied directly to long-form inputs without further fine-tuning. This zero-shot setting is inherently challenging, models optimized for short, information-dense texts often fail to generalize effectively to verbose and loosely structured long documents, where the relevant information may be sparse or scattered.
\newline
To address this, we adopt and extend a sentence ranking approach proposed in our prior work (\citet{Prathamesh_2025}) , which was originally designed to improve LDC efficiency in a supervised setting. In contrast, this work focuses on zero-shot transfer, where no long document training data is used, making the setup more challenging and widely applicable. Rather than modifying model architecture or training it on long inputs, we propose a lightweight, data-driven context reduction technique that selects only the most informative sentences from a document using TF-IDF-based ranking. This transforms a long input into a compact, semantically rich input that aligns better with the model’s original short-text training distribution.

Figure \ref{fig:paragraph} illustrates our core idea where a long document is transformed into a shorter, semantically dense version by selecting only the most informative sentences based on TF-IDF scores (\citet{Qaiser_2018}). These highlighted sentences contain key topical cues, enabling a model trained solely on short texts to make accurate predictions. This content reduction strategy preserves essential information while significantly reducing input size and computation, making it well-suited for zero-shot long document classification.

We explore three strategies for context reduction:

\begin{enumerate}
    \item Fixed Sentence Selection: Selecting the top $k$ sentences (where $k \in \{1, 2, 3, 4, 5\}$) based on their TF-IDF scores. This helps assess how much information is retained when only a few informative sentences are provided.

    \item Percentage-Based Selection: Retaining 25\%, 50\%, or 75\% of the document content. For each percentage level, we experiment with four selection strategies - selecting content from the beginning (First), from the end (Last), randomly positioned content (Random), and top-ranked content based on TF-IDF scores (Ranked).

    \item Score Normalization: TF-IDF scores are normalized to account for differences in sentence lengths and term distributions, improving the fairness and accuracy of sentence ranking and selection.

\end{enumerate}
These techniques are evaluated on the MahaNews dataset, which contains long Marathi news articles labeled by topic (\citet{Mittal_2024,Mirashi_2024}). We fine-tune (\cite{Devlin_2018}) the marathi-bert-v2 model on SHC data (headline-category pairs) and evaluate it on context-reduced LDC data created using the above strategies (\citet{joshi2022l3cubeMahaCorpus}).
Despite significantly reducing the input length, our context-reduction strategies yield classification performance that closely mirrors that of the full-context LDC setup. In particular, selecting only a subset of highly informative sentences, either through a small fixed count or a reduced percentage of the document, maintains accuracy comparable to using the complete document. In certain cases, these context-reduced inputs even outperform the full-context baseline, indicating that a substantial portion of the original content may be redundant or less informative. These results highlight the effectiveness of our method in preserving core semantic information while substantially improving inference efficiency(\cite{Liu_2018b, Moro_2023}).

\subsection{Key Contributions:}
\begin{itemize}
    \item We address the challenge of efficient zero-shot Long Document Classification (LDC) by adapting transformer models (\citet{Park_2022}) trained solely on short documents without requiring any additional training on long-document data.
    \item We build upon our prior sentence ranking approach to propose a TF-IDF-based context reduction strategy that intelligently selects informative sentences from long documents, enabling compatibility with short-input models (\cite{Prathamesh_2025}).
    \item Our experiments evaluate three strategies: selecting a fixed number of top-ranked sentences, retaining a percentage of content using multiple heuristics (First, Last, Random, Ranked), and applying double lambda normalization to balance sentence length and informativeness.
    \item We show that normalized ranked selection consistently achieves near-peak or even higher accuracy compared to full-context inputs, while significantly reducing inference time, demonstrating that accurate classification is achievable with minimal context.
    \item The results demonstrate a scalable, architecture-agnostic approach that enables practical and efficient LDC in real-world scenarios, even when models are trained only on short-text datasets.
\end{itemize}

\section{Related Work}

Long Document Classification (LDC) poses a challenge to traditional transformer models due to their limited input length and computational overhead (\citet{Zaheer_2020}). Existing literature offers two primary lines of solutions: model-based architectural innovations and data-based input optimization techniques.


Model-based approaches aim to extend the model’s capacity to process longer sequences efficiently. Techniques such as sparse attention (\citet{AlQurishi_2022})(e.g., Longformer, BigBird) reduce computational overhead by attending to selected tokens (\citet{Zafrir_2019}),(\citet{Beltagy_2020}), while quantization lowers precision to minimize memory usage, enabling deployment on constrained hardware (\citet{Zaheer_2020}). Other techniques include recurrent mechanisms like LSTMs (\citet{Martins_2020}),(\citet{Putri_2023}) (for sequence retention) (\citet{Wagh_2021, Minaee_2021}),(\citet{Teragawa_2021}) and pre-layer normalization to stabilize training (\citet{Pham_2024}). Though effective, these methods often require substantial architectural changes and increased implementation complexity.

In contrast, data-based approaches restructure inputs without altering the model. Common strategies include document chunking (\citet{Song_2024}), followed by aggregation using hierarchical or ensemble techniques (\citet{Yang_2016}),(\citet{Khandve_2022}), and extractive methods that identify the most informative sentences or paragraphs to reduce input length (\citet{Li_2018, He_2019}). These approaches are lightweight, architecture-agnostic, and particularly beneficial for zero-shot or resource-constrained scenarios, where retraining is not feasible (\citet{Prabhu_2021}).
Our work builds upon this second line of research by proposing a sentence ranking-based context reduction strategy, enabling efficient zero-shot LDC using models trained solely on short documents.

\subsection{Training and Evaluation on Native Long Document Datasets
}
In our earlier work (\citet{Prathamesh_2025}) , we proposed a data-driven strategy specifically designed for Long Document Classification (LDC) by optimizing the input representation for both the training data and the testing data. Rather than modifying the underlying model architecture, the focus was on selecting only the most relevant sentences from a long document for training as well as during the inference, using a TF-IDF-based ranking mechanism (\citet{Liu_2018}). This allowed transformer models, typically built for shorter inputs, to perform well on longer texts.
\newline
\newline
Each sentence in a document was treated as a standalone unit and scored based on the TF-IDF values of the words it contained (\citet{Kim_2019, Das_2020}). Then, the highest-ranking sentences were selected to represent the entire document. Multiple selection strategies were explored: selecting a fixed number of top ranked sentences, extracting a percentage of sentences based on rank, and combining TF-IDF with sentence length normalization to better capture informativeness.
This input reduction led to a significant drop in inference time while retaining competitive classification performance. The approach was evaluated on the MahaNews dataset of long Marathi articles(\citet{Mittal_2024}), demonstrating that selective sentence extraction could preserve key semantic cues necessary for accurate classification. Our results confirmed that TF-IDF-based selection methods outperform simpler strategies like using the first, last, or random sentences.
\newline
\newline

Building on this, the current work explores zero-shot generalization by applying this LDC compression technique to adapt long documents for short headline classification models. This not only reduces computational requirements but also opens new directions for model reuse across varied input lengths and text types (\citet{Bamman_2013}). The ability to apply models trained on short, information-dense inputs to verbose, multilingual documents aligns with recent efforts in Indic language processing, where diverse datasets such as L3Cube-IndicNews(\citet{Mirashi_2024}) and MahaCorpus(\citet{joshi2022l3cubeMahaCorpus}) have enabled broader experimentation. Additionally, the development of multilingual Transformer models (\citet{Jain_2020}) and benchmarking efforts like the Long Range Arena (\citet{Tay_2021}) provide strong foundations for investigating scalable architectures. These advances, combined with practical fine-tuning insights from (\citet{Sun_2020}), support the feasibility and relevance of our zero-shot LDC approach in low-resource and cross-lingual scenarios.

\section{Methodology}
\begin{figure*}[]
   \begin{center}
    \includegraphics[height=10cm]{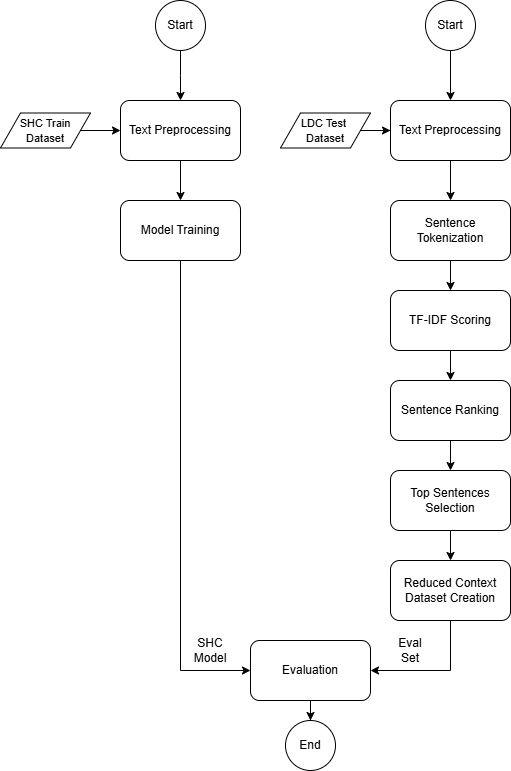}
\end{center}
    \caption{Workflow for SHC-to-LDC Classification using TF-IDF-Based Sentence Selection. The left pipeline shows model training on short headline data, while the right pipeline illustrates the process of sentence tokenization, TF-IDF scoring, ranking, and top sentence selection to create reduced-context inputs for evaluating the SHC model on long documents.
}
    \label{fig:process_diagram}
\end{figure*}

Our goal is to enable effective cross-domain generalization by adapting the marathi-bert-v2 model trained solely on short headlines to effectively classify long documents without modifying the model architecture or repeated training on long document inputs.The proposed methodology follows two parallel yet converging pipelines: one for training the model on SHC data, and the other for evaluating it on LDC inputs after context reduction. This process is depicted in Figure \ref{fig:process_diagram}, which outlines the end-to-end workflow of our short-to-long classification strategy.

\subsection{SHC Training Pipeline}
The training pipeline starts with preprocessing the SHC training dataset from the L3Cube’s IndicNews Marathi corpus. This involves basic text cleaning and stop-words removal (\citet{Das_2023}). The preprocessed headlines, paired with their corresponding topic labels, are then used to fine-tune the marathi-bert-v2 transformer model, which serves as our base classification model. Since SHC inputs are concise and information-dense, the model learns to make accurate predictions using limited but focused context.
\newline\newline
\subsection{LDC Evaluation Pipeline with TF-IDF-Based Reduction}
For long news articles in the LDC test set, a preprocessing pipeline is applied to reduce context while retaining semantically rich content. First, each article is split into individual sentences using the Indic NLP tokenizer. Next, we compute TF-IDF scores for each sentence to quantify its informativeness. Each sentence ($S_i$) is treated as a document in the TF-IDF framework. The term-level TF-IDF score is calculated using:
\[
\text{TF-IDF}(t_i) = \text{TF}(t_i) \cdot \text{IDF}(t_i)
\]
Components:
\newline
Term Frequency (TF): Measures how often a term appears within a sentence, relative to the total number of terms in that sentence. TF indicates the relative significance of a term based on its occurrence within a sentence.
\[
\text{TF}(t_j) = \frac{\text{Frequency\ of\ } t_j \text{\ in\ } S_i}{\text{Total\ number\ of\ terms\ in\ } S_i}
\]
\newline
Inverse Document Frequency (IDF): Captures the uniqueness of a term across the entire set of sentences in a document. Inverse Document Frequency (IDF) reflects how unique or rare a term is across the entire document or corpus.
\[
\text{IDF}(t_j) = \log\left(\frac{N}{1 + \text{Sentence\ frequency\ of\ } t_j}\right)
\]
\newline
Where N is the total number of sentences in the article, and DF($t_j$) is the number of sentences containing the term $t_j$.
\newline\newline
To compute the overall informativeness of a sentence $S_i$, we aggregate the TF-IDF scores of its constituent terms:
\[
\text{Score}(S_i) = \sum_{t_j \in S_i} \text{TF-IDF}(t_j)
\]
\newline\newline
\textbf{Need for Normalization}
\newline\newline
When using the sum of TF-IDF scores to rank sentences, longer sentences naturally tend to receive higher scores, not necessarily because they are more informative, but simply because they contain more terms. This creates an unintended length bias.
\newline
To mitigate this, normalization techniques are employed. One common approach involves dividing the cumulative TF-IDF score by the number of tokens in the sentence. This yields an average TF-IDF score per token, offering a fairer basis for comparing sentences of varying lengths.
\newline
\[
\text{Normalized\_TF\_IDF($S_i$)} = \frac{\text{Score}(S_i)}{\text{Length}( S_i)}
\]
\newline
Where, Length$(S_i)$ refers to the total number of tokens in the sentence.
\newline
However, while this adjustment addresses the preference toward longer sentences, it can introduce an opposite bias: favoring shorter sentences that may have high average TF-IDF scores due to containing a few unique terms. Consequently, important but longer, context-rich sentences might be underrepresented.
\newline\newline
\textbf{Balancing Content Richness and Brevity}
\newline\newline
To counteract this new bias and ensure a more equitable ranking process, a composite scoring strategy is adopted. This method combines normalized TF-IDF with sentence length in a weighted fashion:
\newline

$
\text{Final Normalized Score($S_i$)}=(\lambda_1 \cdot \text{Normalized\_TF\_IDF}(S_i))+(\lambda_2 \cdot \text{Length($S_i$}))
$
\newline\newline
Here:

$\lambda_1 + \lambda_2 = 1$ and $\quad 0 \leq \lambda_1, \lambda_2 \leq 1$
\newline\newline
The parameters $\lambda_1$ and $\lambda_2$ serve as tunable weights that dictate the influence of term importance versus contextual richness:

\begin{itemize}
    \item Higher $\lambda_1$ values emphasize term rarity and uniqueness (favoring high TF-IDF).
    \item Higher $\lambda_2$ values favor longer sentences that may provide more comprehensive context.

\end{itemize}
This hybrid scoring mechanism allows for controlled trade-offs between conciseness and depth, enabling the sentence selection process to be tailored according to the specific goals of the task, such as summarization or training data preparation. The dynamic interplay between TF-IDF and sentence length ensures that both short, impactful sentences and longer, content-rich ones are given appropriate consideration, depending on how the weights are set.

\subsection{Sentence Selection and Reduced Context Creation}

To create reduced-length inputs suitable for a short-context model, we extract the most informative sentences based on their TF-IDF ranks. Sentences are then reassembled in their original order to maintain coherence. Selection is performed in two primary ways:
\newline
\begin{itemize}
    \item \textbf{Fixed Top-N Selection}: We choose the top 1 to 5 ranked sentences incrementally to simulate progressively richer context and study the trade-off between added information and classification performance.

    \item \textbf{Percentage-Based Selection}: We retain 25\%, 50\%, or 75\% of the document content using different heuristics, including selecting the first, last, random, or top-ranked sentences to assess how sentence position and informativeness influence performance.

\end{itemize}
These strategies collectively help us evaluate how much and which parts of the long document are essential for accurate classification while keeping the input compact, mimicking the dense style of SHC inputs.
\newline\newline
\subsection{Evaluation}
The previously trained SHC model is then evaluated on the reduced LDC inputs. We measure classification accuracy and inference time to analyze the effectiveness of the TF-IDF-based reduction in retaining essential information while minimizing input length. Additionally, we compare this ranking approach with other naive methods like selecting either the first, last, or random portions of sentences at various percentages, to evaluate the impact of positional context selection. This multi-pronged approach enables us to assess how different sentence selection and context reduction techniques affect classification accuracy and inference time without altering the underlying model architecture.

\begin{table}[h!]
\fontsize{15}{20}\selectfont
\resizebox{0.5\textwidth}{!}{ 
\begin{tabular}{|c|c|c|c|c|c|}
\hline
\textbf{Sentence(s)} & \textbf{First} & \textbf{Last} & \textbf{Random} & \textbf{Ranked} & \shortstack{\textbf{Ranked} \\ \textbf{Normalized}} \\ \hline
1 & 70.21\% & 68.10\% & 70.11\% & 75.45\% & 73.45\% \\ \hline
2 & 76.02\% & 73.80\% & 70.56\% & 77.64\% & 77.96\%\\ \hline
3 & 75.42\% & 74.93\% & 73.21\% & 76.90\% & 78.40\% \\ \hline
4 & 74.83\% & 74.11\% & 73.54\% & 76.82\% & 79.84\% \\ \hline
5 & 76.12\% & 74.96\% & 74.13\% & 79.84\% & 79.84\%\\ \hline
\end{tabular}
}

\caption{Sentence-wise Accuracy Results — The table shows accuracy across different sentence selection strategies (first, last, random, ranked, ranked normalized) for 1 to 5 selected sentences.}
\label{tab:sentence_comparison}
\end{table}
\begin{figure}
    \includegraphics[width=0.5\textwidth]{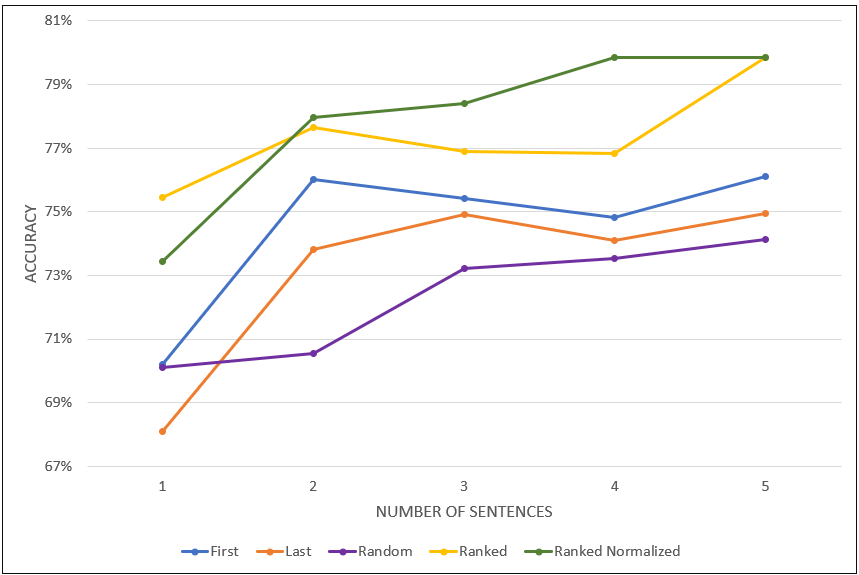}
    \caption{Sentence-wise accuracy graph — The graph visualizes sentence-wise accuracy for different selection methods: first, last, random, and ranked. It plots accuracy against the number of selected sentences (1 to 5). The graph shows that ranked > first > last> random.}
    \label{fig:graph1}
\end{figure}
\section{Results and Discussion}
When trained on the SHC training dataset and evaluated on the full-length LDC dataset, the marathi-bert-v2 model achieves a baseline classification accuracy of 80.314\%. This serves as our reference point for assessing the effectiveness of reduced-context strategies. The goal of our experiments is to reach or closely approach this baseline using significantly shorter inputs. By doing so, we aim to reduce the model’s inference time and computational load, while preserving classification performance.

\subsection{Fixed Number of Sentences Approach}
The results presented in Table \ref{tab:sentence_comparison} show the classification accuracy obtained by selecting a fixed number of sentences (1 to 5) using four different strategies: First, Last, Random, and our TF-IDF-based Ranked method. Figure \ref{fig:graph1} provides a graphical view of the same results, making trends easier to observe. Accuracy consistently improves with more selected sentences across all strategies, with the TF-IDF-based ranking method outperforming others in most cases. For instance, selecting just 2 ranked sentences yields an accuracy of 77.64\%, which surpasses the 5-sentence accuracy of the random (74.13\%) and last (74.96\%) strategies.

This shows that our sentence ranking strategy is effective in prioritizing the most informative content for classification, especially when limited to a small number of input sentences.With just the top 5 ranked sentences, the model reaches 79.84\% accuracy, demonstrating a negligible drop of around 0.6\% from the full-context baseline of 80.314\%, while significantly reducing input size. Since this method uses a fixed number of sentences regardless of document length, it offers a predictable and reduced inference cost, making it both efficient and scalable for real-world applications.

Ranked normalization demonstrates consistently strong performance, with accuracy steadily improving as the number of selected sentences increases. At  2 sentences, it reaches 77.96\%, slightly exceeding Ranked (77.64\%) and clearly ahead of other strategies. By 5 sentences, Ranked and Ranked Normalized both achieve a peak accuracy of 79.84\%, which is remarkably close to the full-context baseline of 80.314\%, reflecting only a 0.47\% drop in performance while using far less input.
The benefit of normalization lies in its ability to fairly compare sentences of varying lengths and densities, enabling better identification of semantically rich content. For example, Ranked Normalized achieves 78.40\% accuracy with 3 sentences, surpassing other approaches at the same level and retaining over 97.7\% of the baseline accuracy with significantly reduced input.

These findings demonstrate that informed sentence selection, particularly when combined with score normalization, can effectively reduce input size without significantly compromising classification performance. This makes the approach highly suitable for scalable long document classification, especially in cross-domain settings where models trained on short texts are applied to much longer inputs.
\newline
\subsection{Percentage Based Approach}
\renewcommand{\arraystretch}{1.5} 
\begin{table}[h!]
\fontsize{15}{20}\selectfont
\centering
\resizebox{0.5\textwidth}{!}{
\begin{tabular}{|c|c|c|c|c|c|}
\hline
\shortstack{\textbf{Percentage} \\ \textbf{Selected}} & \textbf{First} & \textbf{Last} & \textbf{Random} & \textbf{Ranked} & \shortstack{\textbf{Ranked} \\ \textbf{Normalized}} \\ \hline
25\% & 75.65\% & 78.10\% & 77.97\% & 78.31\% & 78.20\% \\ \hline
50\% & 78.08\% & 79.53\% & 79.50\% & 79.56\% & 79.53\% \\ \hline
75\% & 78.75\% & 79.82\% & 79.68\% & 80.20\% & 80.50\% \\ \hline
100\% & 80.23\% & 80.23\% & 81.18\% & 81.29\% & 81.18\% \\ \hline
\end{tabular}
}
\caption{Percentage-wise accuracy table — This table shows the classification accuracy achieved by various sentence selection strategies as the percentage of sentences selected increases in increments of 25\% (i.e.,25\%, 50\%, 75\%, and 100\%).
}
\label{tab:percentage_comparison}
\end{table}

\begin{figure}
    \raggedright 
    \includegraphics[width=0.5\textwidth]{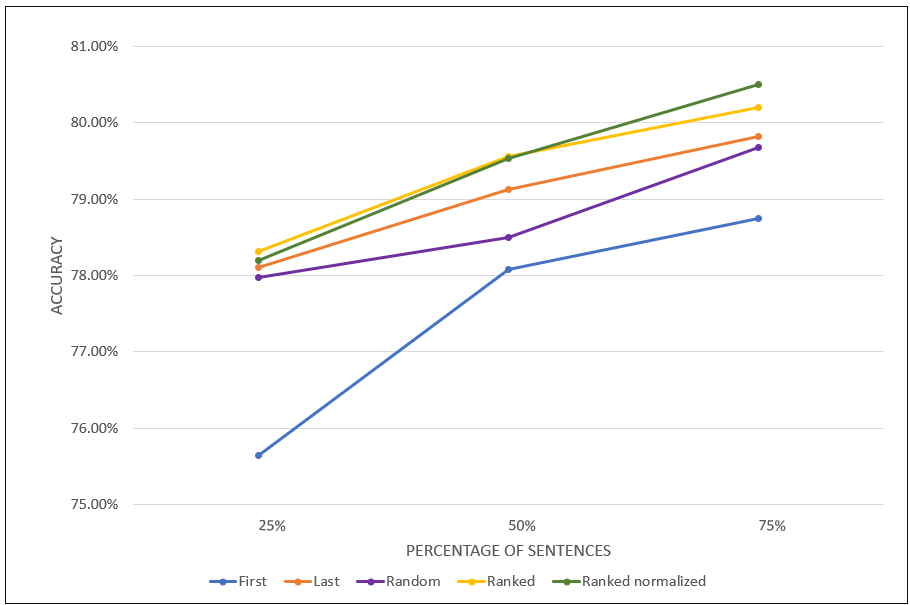}
    \caption{Percentage-wise accuracy graph — This graph shows the classification accuracy achieved by various sentence selection strategies as the percentage of sentences selected increases in increments of 25\% (i.e.,25\%, 50\%, 75\%, and 100\%). The x-axis represents the proportion of selected content, while the y-axis reflects the corresponding accuracy. Each line depicts a method: First, Last, Random, Ranked, and Ranked Normalized.
    }
    \label{fig:graph2}
\end{figure}
Table \ref{tab:percentage_comparison} and Figure \ref{fig:graph2} summarize the classification accuracy achieved using different sentence selection strategies which are First, Last, Random, Ranked, and Ranked Normalized, as the percentage of retained content varies from 25\% to 75\%, with 100\% representing the full-document baseline. A key observation is that the Ranked Selection method consistently performs best across all levels of content reduction. At 25\% retention, it achieves 78.31\% accuracy, already close to the baseline of 80.314\%, and better than all other methods at the same level. This early lead indicates that even a quarter of well-ranked content can retain much of the semantic core of the document.

As we move to 50\% retention, Ranked Normalized and Ranked both reach 79.53\% and 79.56\% respectively, showing minimal degradation of accuracy while cutting input size in half. In contrast, the First, Last, and Random strategies yield lower scores and are more sensitive to what portion of the document is dropped. The graph further reinforces this trend, with the green line (Ranked Normalized) consistently staying on top, followed by yellow (Ranked) and orange (Last). The blue line (First) starts off significantly lower, reflecting the fact that initial sentences may not always capture the most relevant context.

At 75\% retention, Ranked Normalized reaches 80.50\%, surpassing even the full-context baseline of 80.314\%. This indicates that removing the bottom 25\% of low-ranked sentences actually improves classification, likely by eliminating noise. This peak performance is also visible in the graph, where the Ranked Normalized curve continues to rise steadily. Finally, at 100\% content (full document), all methods converge close to the baseline with Ranked Normalized and Ranked hitting 81.18\% and 81.29\% respectively.

These results underscore the strength of TF-IDF-based ranking, particularly when enhanced with normalization to account for sentence length and term density. Not only does this approach preserve classification accuracy across aggressive reductions, but it also occasionally exceeds full-document performance, highlighting its ability to filter noise and focus on semantically rich content. This makes it a robust, resource-efficient pre-processing strategy for scalable long document classification.

\subsection{Inference Time}

One of the core aims of context reduction is to lower the time taken during model inference, all while preserving the quality of classification results. Our strategy leverages TF-IDF-based sentence filtering to dynamically shorten input lengths according to the true content density of each document, instead of using arbitrary maximum token limits (like BERT’s default 512 tokens). Rigid limits often result in unnecessary padding or truncation, which wastes computation or cuts off valuable content. In contrast, our adaptive selection method ensures that the model processes only the most informative segments, leading to faster and smarter inference.

In the sentence-wise approach, we observe a clear advantage in terms of inference efficiency. Selecting a fixed number of TF-IDF–ranked sentences from each document keeps the input size predictable and small. With just 1 sentence, the inference time is as low as 22.5 seconds, gradually increasing to 27.1 seconds for 5 sentences, which is less than half of the evaluation time for the full context long documents. Even at 5 sentences, this method remains faster than any percentage-based approach, making it especially suitable for real-time or low-resource scenarios. The controlled input size ensures that the model operates within computational limits while still accessing the most informative content.

In the percentage-wise approach, inference time increases significantly with the amount of content retained. At 25\%, the time is 30.3 seconds, rising to 41.7 seconds for 50\%, and reaching 51.9 seconds at 75\%. Full-document inference at 100\% takes 57.9 seconds, almost twice as long as 25\%. This trend reflects the direct correlation between input size and computation cost, highlighting the inefficiency of processing entire documents when much of the content may be redundant or less relevant.

A key takeaway from these results is that smart sentence selection strategies, especially the fixed sentence approach using TF-IDF ranking, achieve significant reductions in inference time while maintaining competitive classification accuracy. These techniques make long document classification more scalable and practical by ensuring that only the most essential parts of the text are processed, thereby preserving accuracy and drastically cutting down computation time.

\begin{figure}[t] 
    \centering
    \includegraphics[width=0.5\textwidth]{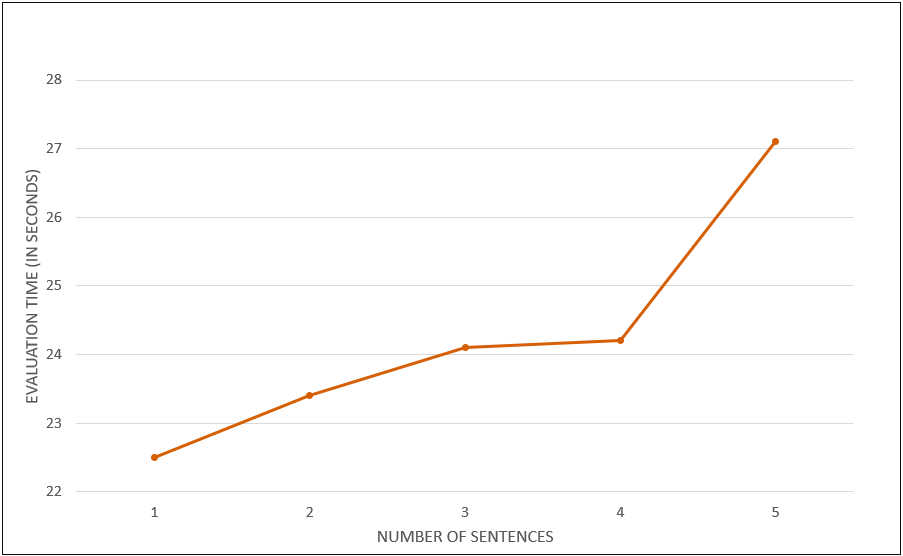}
    \caption{Evaluation time graph for sentence-wise selection — This graph illustrates how evaluation time (in seconds) varies with the number of sentences selected during sentence-wise reduction. The x-axis indicates the count of selected sentences, while the y-axis reflects the time taken for evaluation. The trend observed suggests that as more sentences are included, the evaluation time gradually increases.}
    \label{fig:graph4}
\end{figure}

\begin{figure}[t] 
    \centering
    \includegraphics[width=0.5\textwidth]{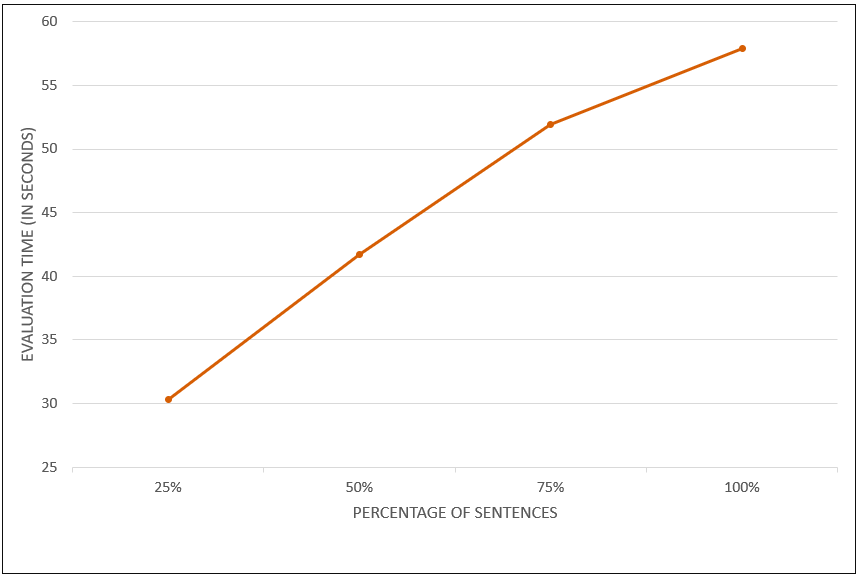}
    \caption{Evaluation time graph for percentage-based selection — This graph displays how evaluation time (in seconds) varies with different percentages of sentence selection. It highlights the trend in processing time as the proportion of selected sentences increases.}
    \label{fig:graph5}
\end{figure}

\section{Conclusion}
We presented a zero shot strategy for extending the capabilities of short-text classification models to long document classification (LDC). By leveraging TF-IDF-based sentence selection and context reduction techniques, our approach enables a model originally trained on Short Headline Classification (SHC) to effectively handle longer, more complex inputs without architectural modifications.
Through experiments on the MahaNews dataset, we demonstrated that reducing input context using sentence-wise and percentage-based ranking methods can preserve accuracy while significantly lowering inference time. Among all strategies, selecting a percentage of top-ranked sentences, especially using TF-IDF score normalization, consistently yielded the highest classification accuracy. At 75\% content retention, the Ranked Normalized approach even surpassed the full-context baseline of 80.314\%, highlighting its effectiveness in balancing performance with reduced computational overhead. 
Notably, at 50\% content retention, the Ranked Normalized strategy achieves a near-baseline accuracy. Despite the minimal accuracy drop, it leads to a notable reduction in evaluation time, from 57.9 seconds at full context to 41.7 seconds, marking a 28\% improvement in computational efficiency. Even more remarkably, when using the top 5 TF-IDF-ranked sentences, the model achieves 79.84\% accuracy, which is almost on par with the full-context performance, while slashing inference time by more than 50\%, highlighting the effectiveness of sentence ranking for efficient long document classification.
Our work highlights the feasibility of adapting SHC-trained models to long-form inputs using lightweight, data-driven techniques. This approach provides a scalable solution for real-world applications where long texts must be processed efficiently without the need for retraining or altering the underlying model.

\section*{Acknowledgement}
This work was undertaken with the mentorship of L3Cube, Pune. We sincerely appreciate the invaluable guidance and consistent encouragement provided by our mentor during this endeavor.








\bibliography{main}
\bibliographystyle{unsrtnat}

\end{document}